\documentclass[sigconf]{acmart}
\AtBeginDocument{%
  }

\setcopyright{cc}
\setcctype{by}
\copyrightyear{2027}
\acmYear{2027}
\acmDOI{10.1145/3862322.3863543}
\acmConference[WSDM '27]
{Proceedings of the Twentieth ACM International Conference on Web Search and Data Mining}
{February 15--19, 2027}
{Hong Kong, Hong Kong}
\acmBooktitle{Proceedings of the Twentieth ACM International Conference on Web Search and Data Mining (WSDM '27), February 15--19, 2027, Hong Kong, Hong Kong}
\acmISBN{979-8-4007-3117-4/2027/02}

\usepackage{xspace}
\usepackage{xcolor}
\usepackage{graphicx}
\usepackage{subcaption}
\usepackage[normalem]{ulem}
\usepackage{amsmath}
\usepackage{enumitem}

\usepackage{booktabs}
\usepackage{multirow}
\usepackage{makecell}
\usepackage{ragged2e}

\providecommand{\up}[1]{\,{\scriptsize\textcolor{green!45!black}{(#1)}}}
\providecommand{\down}[1]{\,{\scriptsize\textcolor{red!70!black}{(#1)}}}

\newcommand{\vanilla}{Vanilla ReAct\xspace}
\newcommand{\hiexp}{HiExp\xspace}
\newcommand{\ace}{ACE\xspace}
\newcommand{\method}{FINDER\xspace}

\newcommand{\qwenm}{Qwen3.6-35B-A3B\xspace}
\newcommand{\gptoss}{GPT-OSS-120B\xspace}
\newcommand{\deepseek}{DeepSeek-V4-Flash\xspace}
\newcommand{\qwens}{Qwen3.5-9B\xspace}

\newcommand{\reflector}{reflector\xspace}
\newcommand{\optimizer}{editor\xspace}

\begin{document}

\title{FINDER: Studying and Repairing What Test-Time Learning Changes in Deep-Research Agents}

\author{Yinxi Li}
\authornote{Both authors contributed equally to this research.}
\affiliation{%
  \institution{University of Waterloo}
  \city{Waterloo}
  \state{Ontario}
  \country{Canada}}
\email{yinxi.li@uwaterloo.ca}

\author{Qingcheng Zeng}
\authornotemark[1]
\affiliation{%
  \institution{Northwestern University}
  \city{Evanston}
  \state{Illinois}
  \country{USA}}
\email{qingchengzeng2027@u.northwestern.edu}

\author{Weihao Xuan}
\affiliation{%
  \institution{The University of Tokyo}
  \city{Tokyo}
  \country{Japan}}
\email{weihaoxuan@g.ecc.u-tokyo.ac.jp}

\author{Heli Qi}
\affiliation{%
  \institution{RIKEN AIP}
  \city{Tokyo}
  \country{Japan}}
\email{qiheli1998@gmail.com}

\author{Fuheng Zhao}
\affiliation{%
  \institution{University of Utah}
  \city{Salt Lake City}
  \state{Utah}
  \country{USA}}
\email{fuheng.zhao@utah.edu}

\author{Puxuan Yu}
\affiliation{%
  \institution{Snowflake Inc.}
  \city{Menlo Park}
  \state{California}
  \country{USA}}
\email{pxyuwhu@gmail.com}

\author{Kaize Ding}
\affiliation{%
  \institution{Northwestern University}
  \city{Evanston}
  \state{Illinois}
  \country{USA}}
\email{kaize.ding@northwestern.edu}

\author{Pengyu Nie}
\authornote{Correspondence: Pengyu Nie and Tianyu Liu.}
\affiliation{%
  \institution{University of Waterloo}
  \city{Waterloo}
  \state{Ontario}
  \country{Canada}}
\email{pynie@uwaterloo.ca}

\author{Tianyu Liu}
\authornotemark[2]
\affiliation{%
  \institution{Yale University}
  \city{New Haven}
  \state{Connecticut}
  \country{USA}}
\email{tianyu.liu@yale.edu}

\renewcommand{\shortauthors}{Li et al.}

\begin{abstract}
Test-time learning (TTL) lets a frozen language-model agent keep improving after deployment by turning its own experience into natural-language lessons read from its context.
Its accuracy gains, however, do not reveal what the learned experience changes;
this question remains under-explored for deep research, where an agent must repeatedly search, verify evidence, and decide when to answer.
We present a process-level study on BrowseComp-Plus: we assign every incorrect trajectory from three frozen backbones a primary failure family and use the same families to classify the behavioral targets of experience learned by two representative TTL methods.
Failures are concentrated but model-specific.
General-purpose TTL raises accuracy without reliably repairing the dominant families, even though the learned artifacts already cover those families.
This \emph{coverage--outcome mismatch} shows that naming a failure in experience is not sufficient to change the action that produces it.
Guided by this finding, we propose \method{}, which organizes experience by the \textsc{Find}, \textsc{Verify}, and \textsc{Decide} stages of the search workflow, diagnoses where a failed rollout diverges from a successful one, grounds a corrective lesson in that contrast and in retriever feedback, and curates it into a bounded playbook.
Across four frozen backbones, \method improves judged accuracy by 3.7 to 19.2 points, and the same diagnostic shows that its gains repair each backbone's own bottlenecks with far fewer regressions than the baselines and without issuing more searches.
Our code is available at \href{https://github.com/larryinx/finder}{github.com/larryinx/finder}.
\end{abstract}

\ccsdesc[500]{Information systems~Information retrieval}
\ccsdesc[500]{Computing methodologies~Natural language processing}
\keywords{deep-research agents, test-time learning, agentic search}

\maketitle

\section{Introduction}

Large language models (LLMs) are increasingly deployed as agents that reason, call tools, and act over many steps~\citep{openai2024gpt4technicalreport,Guo_2025,yang2025qwen3technicalreport,yao2023react}.
Deep research is among the most demanding of these applications: to answer a question that identifies its target only through indirect clues, an agent must repeatedly search, connect evidence across sources, and decide when the accumulated support is sufficient to answer~\citep{wei2025browsecompsimplechallengingbenchmark,chen-etal-2026-browsecomp,lu2026monolingualdeepresearchevaluating}.
Recent progress on this task has come mainly from training the agent, whether by supervised fine-tuning on synthesized search trajectories~\citep{du2026openseekerdemocratizingfrontiersearch,li2026openresearcherfullyopenpipeline} or by reinforcement learning against a search environment, often on synthesized questions~\citep{jin2025searchr1trainingllmsreason,thakur2026orbitscalableverifiabledata,tongyideepresearchteam2026tongyideepresearchtechnicalreport,chu2026redsearcherscalablecostefficientframework}.
Conventional post-training typically requires weight updates and substantial rollout compute and occurs before deployment; the deployed agent's own trajectories can nevertheless continue to expose repeated query, verification, and stopping errors.

Test-time learning (TTL) offers a complementary route in which the agent learns from the tasks it encounters: the experience of each task is written into an adaptive state that later decisions read, so the agent keeps improving after deployment~\citep{xuan2026learning}.
That state can enter the model at different levels, as an update to its parameters~\citep{sun2020testtimetrainingselfsupervision,hu2025testtimelearninglargelanguage}, as internal representations such as a latent memory~\citep{wang2024memoryllmselfupdatablelarge}, or as text in its context.
We focus on the context level, which requires no access to the weights and leaves an inspectable artifact.
In context-level TTL, the agent distills the knowledge of what worked and what failed in earlier tasks into natural-language lessons, stores the lessons in an external memory or playbook, and reads that artifact when answering later tasks~\citep{shinn2023reflexionlanguageagentsverbal,zhao2024expelllmagentsexperiential, suzgun-etal-2026-dynamic,zhang2026agenticcontextengineeringevolving, du2025accelerating,ouyang2026reasoningbankscalingagentselfevolving,tang2025agentkbleveragingcrossdomain}.
For example, after failing on several questions that describe an unnamed person through a bundle of ordinary attributes, a search agent may learn to lead its query with the single most distinctive clue, such as an unusual degree or a firm name, rather than restate every clue. Throughout this paper, TTL refers to this context-level setting.
Context-level TTL has been demonstrated mostly on mathematical reasoning, question answering, coding, and general agent benchmarks.
Only a few works bring learned experience into search or deep research~\citep{hao2026stochasticexplorationmakestraining,zhou2025mementofinetuningllmagents}, and they primarily emphasize final-answer accuracy.
Whether the learned experience changes the search decisions that actually fail remains largely unexamined.

\begin{figure*}[t]
  \centering
  \includegraphics[width=0.86\textwidth]{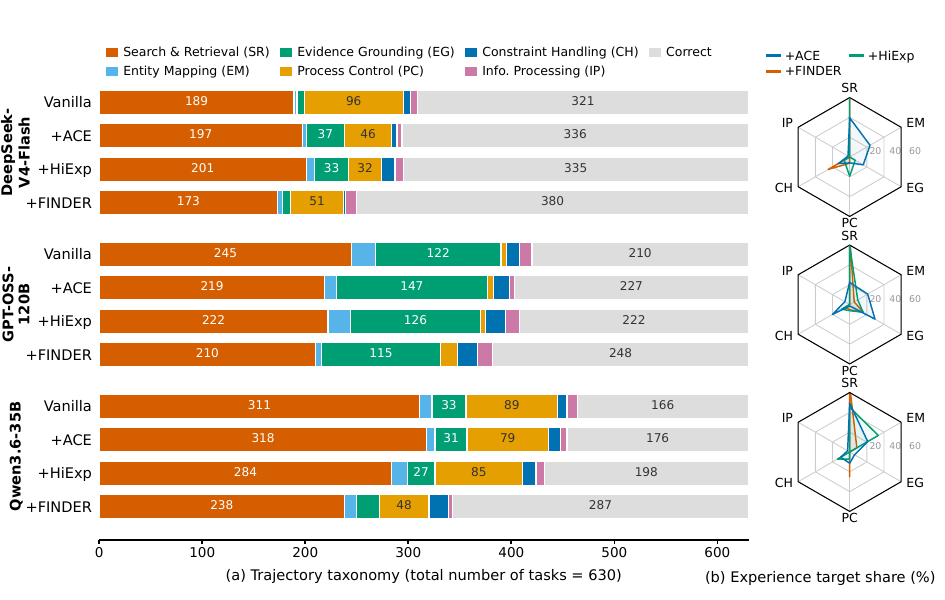}
  \vspace{-5pt}
 \caption{Process-level outcomes and experience targets for three frozen
 search agents. (a) Primary failure family of each incorrect trajectory on
 630 held-out BrowseComp-Plus tasks under \vanilla{}, \ace{}, \hiexp{}, and
 \method{}; gray denotes judged-correct tasks. (b) Within each learned
 artifact, the share of entries whose behavioral target is assigned to each
 failure family. Similar target emphases can accompany substantially different
 changes in the corresponding trajectory failures.
  }
  \Description{Left: horizontal stacked bar charts for DeepSeek-V4-Flash,
  GPT-OSS-120B, and Qwen3.6-35B, each with bars for Vanilla, +ACE, +HiExp,
  and +FINDER that split 630 tasks into six failure families and a gray
  correct segment; search and retrieval is the largest failure segment in
  every bar, and +FINDER has the largest correct segment for each backbone.
  Right: one hexagonal radar chart per backbone showing the percentage of
  playbook entries that target each failure family for ACE, HiExp, and
  FINDER.}
  \label{fig:failure-taxonomy}
  \vspace{-5pt}
\end{figure*}

We therefore begin with a process-level study of what TTL changes in a deep-research agent.
The study holds the environment constant with BrowseComp-Plus~\citep{chen-etal-2026-browsecomp}, whose fixed corpus and shared retriever make search behavior comparable across methods, and runs three frozen backbones with and without the experience learned by two representative TTL methods, Agentic Context Engineering (\ace{})~\citep{zhang2026agenticcontextengineeringevolving} and prompt-based Hierarchical Experience (\hiexp{})~\citep{hao2026stochasticexplorationmakestraining}. To see what changes, we apply one taxonomy from both sides: every incorrect trajectory receives a primary failure family adapted from a deep-research error taxonomy~\citep{wang2026deepresearchagentswrongspanlevel}, and every learned experience entry receives the family it targets.
This two-sided view yields three findings. \textbf{First}, failures are concentrated but model-specific. For \qwenm{}, search and retrieval together with process control account for 86\% of incorrect trajectories, whereas \gptoss{} pairs retrieval with a large evidence-grounding bottleneck. \textbf{Second}, general-purpose TTL raises accuracy without reliably repairing those bottlenecks, and it can displace a failure into another family. \textbf{Third}, the learned artifacts already concentrate on the dominant families, so the gap is not one of topical coverage. We call this discrepancy the \emph{coverage--outcome mismatch}: naming the relevant failure in experience is not sufficient to change the action that produces it.

The mismatch motivates a sharper design hypothesis: useful search experience should identify the decision state in which a lesson applies, test proposed actions against the deployed retrieval environment, and retain lessons in a bounded, actionable structure. We instantiate this hypothesis as \method{}, a weight-frozen TTL framework organized around the three stages of the deep-research workflow: \textsc{Find} sources from the question's clues, \textsc{Verify} them against those clues, and \textsc{Decide} whether the evidence supports an answer. Its playbook stores each lesson under its applicable stage. Its adaptation loop \emph{diagnoses} the stage at which a failed rollout diverges from a successful rollout of the same question, \emph{grounds} a corrective lesson in that contrast and, for query lessons, in feedback from the retriever, and \emph{curates} the lesson into the playbook through a budgeted editor that can add, consolidate, or delete entries.

Across four frozen backbones on BrowseComp-Plus, \method{} improves judged accuracy by 3.7 to 19.2 points and is the strongest evaluated system on every backbone. Returning to the study's diagnostic, we find that its gains land on each backbone's bottlenecks without the error transfer seen under the baselines, that they \mbox{consist} of more repairs in concentrated failure families, and that they are not obtained through more searches. The learned playbooks are small, state-conditioned, and backbone-specific, and reading them with the baselines' artifacts points to a general lesson for TTL in deep research: a large part of what a search agent must learn at test time is how its own search tool behaves, so effective experience does not merely name a failure but records, for the decision state in which the agent breaks down, an action that works with the deployed retriever. In summary, our contributions are:
\begin{itemize}
  \item We give a process-level study of weight-frozen test-time learning for deep research agents, linking held-out trajectory failures to the behavioral targets of learned experience and identifying the coverage--outcome mismatch.
  \item We propose \method{}, which turns recoverable success--failure contrasts into stage-specific, tool-assisted lessons and maintains them with bounded, non-monotonic playbook editing.
  \item We evaluate four frozen backbones and complement accuracy with paired repair/regression counts, failure-family changes, artifact audits, and search productivity, showing both where \method{} helps and where its evidence and scope remain limited.
\end{itemize}

\section{Related Work}

\paragraph{Deep-research agents}
Deep-research agents extend LLM reasoning with iterative search and evidence synthesis over long horizons~\citep{li2025webthinkerempoweringlargereasoning,wei2025browsecompsimplechallengingbenchmark}, and most recent progress is training-centric.
Search-R1 formulates search as a reinforcement-learning environment and trains the model to interleave reasoning with search calls under outcome rewards~\citep{jin2025searchr1trainingllmsreason}.
Later work scales the data side: OpenSeeker synthesizes fact-grounded questions and denoised trajectories for supervised fine-tuning~\citep{du2026openseekerdemocratizingfrontiersearch}, OpenResearcher builds a reproducible offline browsing environment for long-horizon trajectory synthesis~\citep{li2026openresearcherfullyopenpipeline}, and ORBIT generates verifiable multi-hop questions for reinforcement learning on a tight budget~\citep{thakur2026orbitscalableverifiabledata}; end-to-end systems such as Tongyi DeepResearch and REDSearcher combine synthetic data, mid-training, supervised fine-tuning, reinforcement learning, and custom environments~\citep{tongyideepresearchteam2026tongyideepresearchtechnicalreport,chu2026redsearcherscalablecostefficientframework}.
Two studies bring experience into this setting: \hiexp distills hierarchical experience from contrastive analysis of rollouts and multi-level clustering, to guide inference and regularize reinforcement learning~\citep{hao2026stochasticexplorationmakestraining}, and Memento adapts a deep-research agent through case-based memory without fine-tuning~\citep{zhou2025mementofinetuningllmagents}.
Both primarily report accuracy.
We keep the search agent frozen throughout, hold the corpus and retriever fixed through BrowseComp-Plus~\citep{chen-etal-2026-browsecomp}, and examine how learned experience changes the agent's search process rather than only its score.

\paragraph{Experience-based test-time learning}
Experience-based test-time learning at context level lets a frozen LLM accumulate natural-language lessons from its own inference-time interactions and reuse them on later tasks.
Reflexion first showed that verbal reflections kept in an episodic buffer improve subsequent attempts at the same task~\citep{shinn2023reflexionlanguageagentsverbal}, and ExpeL extended this from within-task retries to insights extracted across a pool of past tasks~\citep{zhao2024expelllmagentsexperiential}.
Dynamic Cheatsheet framed the accumulated artifact as a self-curated test-time memory of strategies and reusable snippets~\citep{suzgun-etal-2026-dynamic}.
Agentic Context Engineering (\ace{}) treats the artifact as an evolving playbook and replaces wholesale rewriting with structured incremental updates to avoid brevity bias and context collapse~\citep{zhang2026agenticcontextengineeringevolving}.
ReasoningBank moves to streaming task sequences, distills strategies from both successful and failed experiences, and couples memory with test-time scaling~\citep{ouyang2026reasoningbankscalingagentselfevolving}, and Agent KB shares such experience across heterogeneous agent frameworks through a knowledge base queried at planning and feedback time~\citep{tang2025agentkbleveragingcrossdomain}.
This line has been validated on mathematical reasoning, question answering, coding, finance, and general agent benchmarks, and evaluated primarily by final task outcomes.
We evaluate two representative experience-based methods, \ace{} and \hiexp{}, on deep research and ask a process-level question: which search failures does the learned experience change, and does what it covers align with what it repairs?

\begin{figure*}[t]
  \centering
  \includegraphics[width=0.92\textwidth]{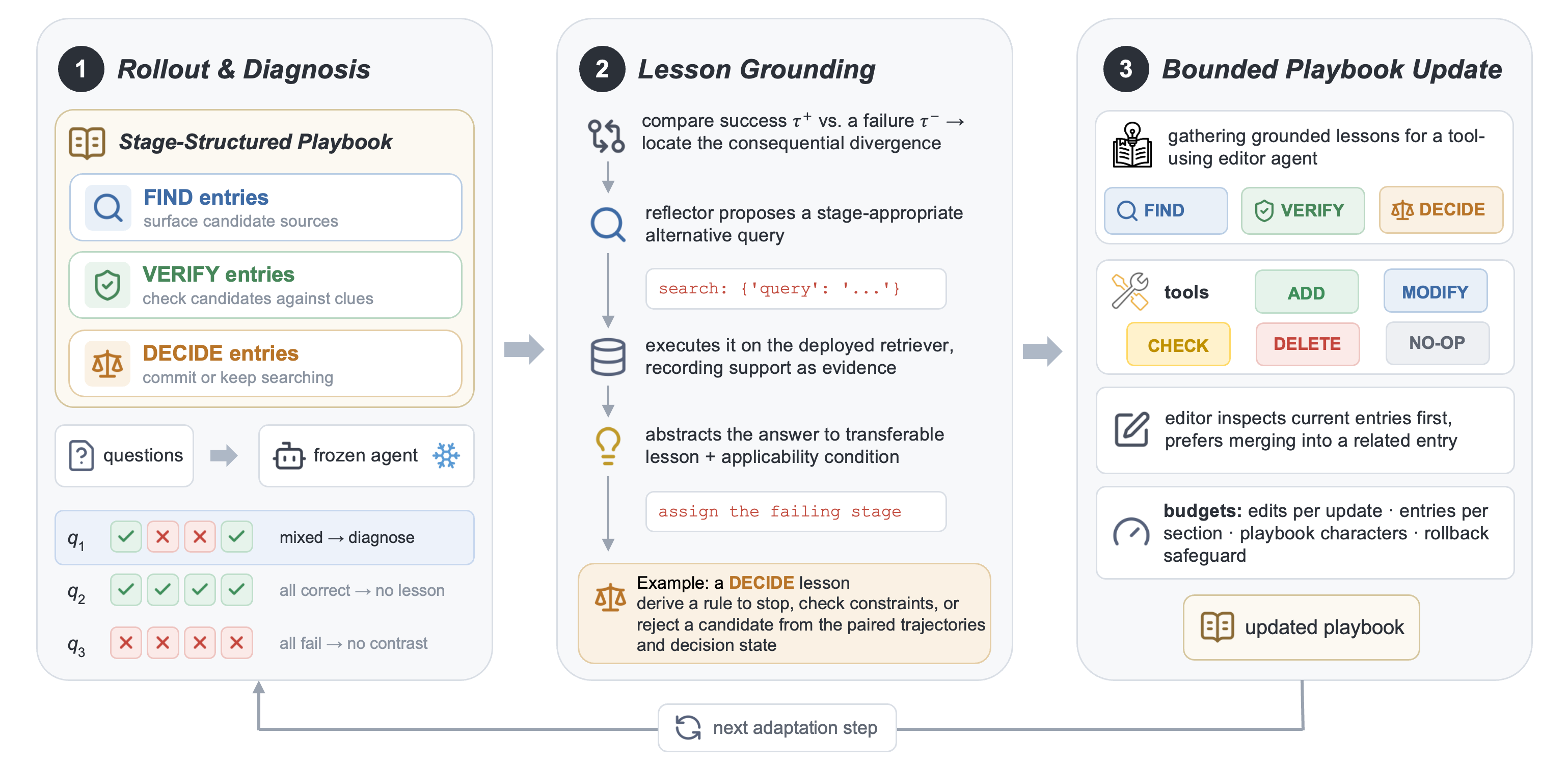}
  \vspace{-8pt}
 \caption{Overview of \method{}. Each adaptation step (1) rolls out
 grouped trajectories under the current stage-structured playbook and
 diagnoses the failures in mixed-outcome groups; (2) grounds a
 stage-specific lesson by contrasting each failure with the most
 efficient success, executing candidate queries on the deployed
 retriever, and abstracting the result into a transferable lesson; and
 (3) curates the lesson into the corresponding playbook section through
 a budgeted tool-using editor. The updated playbook conditions the next
 adaptation step; the agent's parameters remain frozen throughout.}
  \Description{A three-panel pipeline diagram. Panel 1, Rollout and
  Diagnosis, shows a playbook with FIND, VERIFY, and DECIDE entries
  conditioning a frozen agent; questions whose rollouts mix correct and
  incorrect answers are diagnosed, while all-correct and all-failed
  questions are skipped. Panel 2, Lesson Grounding, compares a success with
  a failure, runs a proposed query on the deployed retriever, and abstracts
  the result into a stage-tagged lesson. Panel 3, Bounded Playbook Update,
  shows an editor agent with add, modify, check, delete, and no-op tools
  under edit and size budgets, producing an updated playbook that loops
  back to the next adaptation step.}
  \label{fig:finder-overview}
  \vspace{-5pt}
\end{figure*}

\section{What Does Test-Time Learning Change in a Deep-Research Agent?}
\label{sec:motivation}

We study two representative context-level TTL methods. \ace incrementally curates reflections into a structured playbook, whereas \hiexp contrasts successful and failed rollouts to build hierarchical experience, placing broad strategies in the prompt and retrieving case-level guidance during search~\citep{zhang2026agenticcontextengineeringevolving,hao2026stochasticexplorationmakestraining}. Both specify how experience is constructed and delivered, but they do not guarantee that it changes the domain-specific capabilities on which deep research depends. Deep research makes this distinction observable because each trajectory exposes a sequence of operational decisions: finding candidate sources, verifying them against the question's clues, and deciding whether the evidence supports an answer.
Process-level studies likewise find that errors cluster at identifiable points in this workflow~\citep{wang2026deepresearchagentswrongspanlevel}. We therefore ask: where does a frozen deep-research agent fail (\textbf{RQ1}); which failures do general TTL methods change (\textbf{RQ2}); and does the targeted failure family align with observed failure changes (\textbf{RQ3})?

We answer these questions on BrowseComp-Plus~\citep{chen-etal-2026-browsecomp}, whose fixed corpus and shared retriever hold the search environment constant. We evaluate three backbone models on the same 630 held-out questions, either without learned experience (\vanilla{}) or with our weight-frozen implementations of ACE and prompt-based HiExp. Figure~\ref{fig:failure-taxonomy} additionally reports \method{} as a common taxonomy reference for the later analysis in \S\ref{sec:analysis}. For each incorrect trajectory, we assign one primary cause from six families adapted from the taxonomy of~\citet{wang2026deepresearchagentswrongspanlevel}: search and retrieval (SR), entity mapping (EM), evidence grounding (EG), process control (PC), constraint handling (CH), and information processing (IP). We independently classify the behavioral target of every learned experience entry using the same six families. Full definitions and the annotation protocol appear in Appendix~\ref{sec:appendix:taxonomy}, and the adaptation and evaluation settings appear in \S\ref{sec:experiments}.

\paragraph{Finding 1: search failures form concentrated, model-specific bottlenecks}
The \vanilla{} agent's rows in Figure~\ref{fig:failure-taxonomy} are highly structured, but not identical across backbones. \qwenm{} fails most often in SR (311 of 630 tasks) and PC (89); together these two families explain 86\% of its incorrect trajectories. \deepseek{} has the same two bottlenecks, with 189 SR and 96 PC failures. \gptoss{} instead combines 245 SR failures with a large EG bottleneck (122).
Thus, the shared primary bottleneck is search and retrieval, while the secondary bottleneck depends on the model.
An effective adaptation method should therefore repair the observed workflow bottleneck rather than merely improve aggregate accuracy.

\paragraph{Finding 2: generic experience raises accuracy without reliably repairing those bottlenecks}
For \qwenm{}, \ace{} and prompt-based \hiexp{} raise judged accuracy from 26.3\% to 27.9\% and 31.4\%, respectively. The dominant failures move much less. SR incidence is 49.4\% under \vanilla{}, 50.5\% under \ace{}, and 45.1\% under \hiexp{}; PC incidence changes from 14.1\% to 12.5\% and 13.5\%. Cross-model comparisons make the distinction sharper. \ace{} reduces GPT-OSS's SR count from 245 to 219 while EG rises from 122 to 147. On DeepSeek, \ace{} and \hiexp{} reduce PC from 96 to 46 and 32, but SR rises from 189 to 197 and 201 and EG rises from 7 to 37 and 33. General-purpose TTL can therefore fix, preserve, or displace an error, but a positive net score does not establish that the main search bottleneck was repaired.

\paragraph{Finding 3: the learned experience already covers the relevant topic}
A natural explanation for Finding~2 is that the learned experience simply ignores the dominant failure families. Panel~(b) of Figure~\ref{fig:failure-taxonomy} rules out this simple account. SR is the largest target family in five of the six \ace and \hiexp artifacts. The exception is GPT-OSS under \ace{}, whose largest share targets EG (29.5\%), that model's secondary bottleneck. The mapping from target share to outcome is nevertheless inconsistent. On Qwen, \ace and \hiexp allocate comparable portions of their learned artifacts to SR, with 61 of 124 entries (49.2\%) for \ace and 19 of 42 entries (45.2\%) for \hiexp{}. Despite this similar coverage, SR failures increase from 311 to 318 under \ace{}, while \hiexp reduces them only to 284. The discrepancy also appears across backbones: 67.7\% of DeepSeek's \hiexp entries target SR, while SR failures rise from 189 to 201; GPT-OSS's \ace artifact devotes its largest share to EG, while EG failures rise from 122 to 147. We call this aggregate inconsistency the \emph{coverage--outcome mismatch}. Coarse topical alignment is not sufficient to explain whether learned experience changes behavior, and the audit neither shows that a particular entry was invoked nor identifies why an aligned artifact succeeds or fails. Studying TTL therefore requires connecting artifact targets to observed trajectory changes rather than inferring effectiveness from artifact contents alone.

\paragraph{Implications of the motivation study for \method{}}
The coverage--outcome mismatch motivates state-conditioned, retriever-aware lessons under a bounded memory budget. It does not establish that artifact length or any single design choice causes the observed failures. We instantiate this hypothesis as \method{} and evaluate it with the same diagnostic.

\section{FINDER: Find, Verify, and Decide for Deep-Research Agents}
\label{sec:methodology}

Guided by the motivation study, we introduce \method{}, a weight-frozen TTL framework that turns recoverable failures into stage-structured experience for deep-research agents (Figure~\ref{fig:finder-overview}). Its playbook associates each lesson with a workflow stage (\textsc{Find}, \textsc{Verify}, or \textsc{Decide}) and an applicable decision state. The adaptation loop \emph{diagnoses} where a failed trajectory diverges from a successful one, \emph{grounds} a corrective lesson in this contrast and, when appropriate, retriever feedback, and \emph{curates} it into the corresponding stage section under explicit budgets.

Let $\pi_{\theta}$ denote the search agent with fixed parameters $\theta$. At adaptation step $t$, it is conditioned on an external playbook
\[
  P_t = S \oplus E_t,
\]
where $S$ is a fixed stage-structured skeleton, $E_t$ is the set of learned entries, and $\oplus$ denotes their composition in the agent's context. Given a batch of $B$ questions, \method samples $G$ trajectories for each question and obtains an outcome judgment for every trajectory:
\[
  \begin{aligned}
    \tau_i^g &\sim \pi_{\theta}(\,\cdot\mid q_i,P_t), \\
    r_i^g &= J(q_i,y_i,\tau_i^g) \in \{0,1\}.
  \end{aligned}
\]
Here, $y_i$ denotes the supervision available during adaptation and $J$ is the outcome judge. The rollout agent observes $q_i$ and $P_t$ but not $y_i$, while the judge and \reflector can use $y_i$ to determine the outcome and diagnose a failed rollout. We first describe the stage-structured playbook (\S\ref{sec:methodology:playbook}), followed by contrastive diagnosis (\S\ref{sec:methodology:diagnose}), tool-assisted lesson construction (\S\ref{sec:methodology:ground}), and bounded curation (\S\ref{sec:methodology:curate}). \textsc{Find}, \textsc{Verify}, and \textsc{Decide} are the stages of the deep-research workflow, and this structure attributes each playbook lesson to a core capability of the search agent. It does not introduce a hierarchy into TTL: adaptation remains a single \emph{diagnose--ground--curate} loop over the stage-structured playbook.

\subsection{A Stage-Structured Playbook}
\label{sec:methodology:playbook}

The playbook provides the domain-specific organizing principle: rather than arranging experience by task type or abstraction level, it associates each lesson with the stage of the deep-research workflow in which the corresponding core capability is exercised. A \textsc{Find} entry describes the state before a viable candidate is known and records a discovery action with its rationale and applicability boundary. A \textsc{Verify} entry begins from a candidate or bridge entity and records how to test one relevant attribute or make a multi-hop pivot. A \textsc{Decide} entry instead stores a rule for acting on accumulated evidence, such as when to stop, which constraints must be checked, or when an answer should be withheld.

The fixed skeleton $S$ defines the three stages and gives high-level conventions for interacting with the deployed retriever, while $E_t$ supplies lessons learned from trajectories. Together, the section and entry schema connect an observable decision state to a concrete action and an applicability boundary. To encourage transfer beyond the source question, lessons express actions through entity types or placeholders rather than copying an answer. The complete $P_t$ is placed in the agent's context for every rollout, so stage structure organizes attribution and curation rather than dynamically routing entries at particular turns. Consequently, guidance for all three stages remains available throughout the multi-step trajectory, while each diagnosed failure still has a stable destination in the playbook.

\subsection{The Diagnose--Ground--Curate Loop}
\label{sec:methodology:loop-overview}

Each adaptation step updates the playbook from the grouped rollouts of \S\ref{sec:methodology} through three mechanisms. \emph{Diagnosing} identifies recoverable success--failure contrasts and localizes each failure to a workflow stage, yielding a candidate lesson. \emph{Grounding} tests proposed queries on the agent's own retriever for \textsc{Find} and \textsc{Verify} lessons, while anchoring \textsc{Decide} rules in the observed trajectory pair. \emph{Curating} decides whether and how to incorporate the grounded lesson into its stage section under explicit edit and size budgets.

\subsubsection{Diagnosing the Failing Stage}
\label{sec:methodology:diagnose}

A failed trajectory alone does not establish that a better action lies within the frozen agent's current capabilities, so \method restricts diagnosis to failures with an observable successful alternative. For the $G$ rollouts of each question, it forms the mixed-outcome set $\mathcal{M}_t=\{i:0<\sum_{g=1}^{G}r_i^g<G\}$, in which at least one rollout succeeds and one fails under the same agent, tools, and playbook snapshot. Correct rollouts require no correction, and all-fail groups yield no lesson because they contain no successful peer. Within each group, \method selects the correct trajectory $\tau_i^{+}$ with the fewest search calls, breaking ties by length, and a \reflector compares every failed trajectory $\tau_i^{-}$ with $\tau_i^{+}$, identifies the consequential divergence, and assigns it to one workflow stage: a \textsc{Find} failure never surfaces a viable candidate, a \textsc{Verify} failure mishandles a known candidate or misses a necessary multi-hop pivot, and a \textsc{Decide} failure occurs after answer-bearing evidence is available, when the agent drifts to a look-alike, violates an answer constraint, or never commits an answer. The result is a candidate lesson $e=(h,z,a,\varepsilon,c)$: a trigger $h$, a stage $z\in\{\textsc{Find},\textsc{Verify},\textsc{Decide}\}$, a corrective action $a$, supporting evidence $\varepsilon$, and an applicability boundary $c$. Because the artifacts do not retain a per-step census of all-correct, mixed-outcome, and all-fail groups, we treat the gate's empirical coverage as an unmeasured boundary.

\subsubsection{Tool-Assisted Lesson Grounding}
\label{sec:methodology:ground}

Diagnosis identifies where behavior diverges, and grounding constructs the lesson so that its action is compatible with the deployed search environment. For \textsc{Find} and \textsc{Verify} lessons, the \reflector{} holds the same retriever as the rollout agent: it proposes an alternative query, executes it, records what is retrieved, and abstracts the result into a transferable query pattern with an applicability condition, so the stored guidance is tied to the retrieval interface on which the agent actually operates and its origin remains auditable. For \textsc{Decide} lessons, answer-bearing evidence is already present and another query would not address the error. The correction is instead a decision rule about when to stop searching, which constraints to check before answering, or which candidate to reject. Thus, query actions are grounded through interaction with the search environment, whereas decision rules are grounded in the observed behavior.

\begin{table*}[t!]
\centering
\small
\caption{Judged accuracy (\%) on the 630 held-out BrowseComp-Plus
questions.  Parenthesized values are differences from \vanilla{}; the best
result per backbone is in bold.
}
\vspace{-5pt}
\begin{tabular}{lcccc}
\toprule
& DeepSeek-V4-Flash & GPT-OSS-120B & Qwen3.6-35B-A3B & Qwen3.5-9B \\
\midrule
\vanilla{} & 50.95 & 33.33 & 26.35 & 40.48 \\
+\ace{}    & 53.42\up{$+$2.47} & 36.03\up{$+$2.70} & 28.03\up{$+$1.68} & 39.37\down{$-$1.11} \\
+\hiexp{}  & 53.17\up{$+$2.22} & 35.24\up{$+$1.91} & 31.43\up{$+$5.08} & 39.68\down{$-$0.79} \\
\midrule
+\method{} & \textbf{60.32}\up{$+$9.37} & \textbf{39.37}\up{$+$6.04} & \textbf{45.56}\up{$+$19.21} & \textbf{44.13}\up{$+$3.65} \\
\bottomrule
\end{tabular}

\label{tab:main-results}
\vspace{-5pt}
\end{table*}

\begin{table}[t]
\centering
\small
\caption{Stage ablations of \method{}'s learned playbook on BrowseComp-Plus (judged accuracy, \%; parenthesized differences from full \method{}).
}
\vspace{-5pt}
\begin{tabular}{lcc}
\toprule
& GPT-OSS-120B & Qwen3.6-35B-A3B \\
\midrule
+\method{} (full) & \textbf{39.37} & \textbf{45.56} \\
\midrule
\quad w/o \textsc{Find}   & 37.46\down{$-$1.91} & 41.90\down{$-$3.66} \\
\quad w/o \textsc{Verify} & 38.57\down{$-$0.80} & 43.02\down{$-$2.54} \\
\quad w/o \textsc{Decide} & 37.94\down{$-$1.43} & 42.70\down{$-$2.86} \\
\quad w/o \textsc{F}+\textsc{V} & 38.10\down{$-$1.27} & 42.86\down{$-$2.70} \\
\quad w/o \textsc{F}+\textsc{D} & 37.14\down{$-$2.23} & 40.32\down{$-$5.24} \\
\quad w/o \textsc{V}+\textsc{D} & 36.98\down{$-$2.39} & 40.16\down{$-$5.40} \\
\bottomrule
\end{tabular}

\label{tab:stage-ablation}
\vspace{-10pt}
\end{table}

\subsubsection{Curating the Playbook under Budgets}
\label{sec:methodology:curate}

A grounded lesson need not become a new entry. It may duplicate an existing rule, refine that rule's trigger, conflict with prior experience, or be too narrow to justify its context cost. \method therefore delegates the update to a tool-using \optimizer{} that sees the live playbook: it can list and search entries, inspect them, and choose among \texttt{ADD}, \texttt{MODIFY}, \texttt{DELETE}, or no change, with the diagnosed stage fixing the target section. Writing $\mathcal{L}_t$ for the candidate lessons of the current batch, the update is $E_{t+1}=\mathop{\mathrm{Curate}}(E_t,\mathcal{L}_t;\mathbf{b})$, where the budget $\mathbf{b}$ bounds each edit type, the total edits per update, the entries per section, and the overall playbook length. The \optimizer is instructed to search the target section for related triggers and to prefer \texttt{MODIFY} when a lesson can be absorbed by an existing entry. Operations that exceed an edit or section limit are refused. Additions and modifications are also checked against a global length cap, with a final rollback safeguard if an edit pass exceeds the hard limit. These constraints make consolidation and deletion first-class update operations rather than treating experience as a monotonically growing log.

\subsection{Closing the Adaptation Loop}
\label{sec:methodology:loop}

At each adaptation step, \method generates all grouped rollouts under the same playbook snapshot $P_t$, diagnoses and constructs candidate lessons, and curates them into $E_{t+1}$. The next batch is then generated under $P_{t+1}=S\oplus E_{t+1}$. Evidence for each update therefore comes from behavior elicited by the current playbook, allowing later lessons to respond to the effects of earlier revisions. Model parameters remain fixed throughout, while the external playbook is the only persistent adaptation state.

Together, the stage-structured playbook and the iterative sequence of diagnosis, tool-assisted lesson construction, and curation operationalize the design hypothesis developed in \S\ref{sec:motivation}: experience should provide stage-specific, environment-aware guidance for the behaviors that produce search failures. Because the search agent remains frozen, \method isolates the evolving playbook as the object of adaptation. This design enables the study to evaluate final accuracy, trace how the learned artifact evolves, and test whether the workflow failures it targets become less frequent.

\section{Experimental Setup and Results}
\label{sec:experiments}

We evaluate all methods on BrowseComp-Plus~\citep{chen-etal-2026-browsecomp} using 100 questions for adaptation and a disjoint set of 630 questions for held-out evaluation. Each adaptive method learns only from the adaptation set, and we freeze its selected experience artifact before held-out evaluation.
We study four frozen backbone models spanning different families and levels of baseline performance: \deepseek{}, \gptoss{}, \qwenm{}, and \qwens{}. For each backbone, we compare \vanilla without learned experience, our instantiation of \ace{} with gold answers available during reflection, our prompt-based implementation of \hiexp{} with hierarchical experience distilled from success--failure rollout pairs, and \method{}.
All systems share a common ReAct agent loop, question template, document corpus, and dense retriever.
During held-out evaluation, each trajectory is limited to 50 search calls and 50 interaction turns.
Correctness is determined by a common LLM judge using the official BrowseComp-Plus grading prompt, and we report judged accuracy.
The failure-annotation protocol used for the process-level analyses in \S\ref{sec:motivation} and \S\ref{sec:analysis} is described in Appendix~\ref{sec:appendix:taxonomy}.
Appendix~\ref{sec:appendix:settings} presents the remaining experimental details, and \S\ref{sec:appendix:baselines} provides reproduction details of the two baselines.

\subsection{Main Results}
\label{sec:experiments:main}

Table~\ref{tab:main-results} shows that \method{} consistently improves frozen search agents across model families. It raises accuracy over \vanilla{} by $3.7$ to $19.2$ points and achieves the highest accuracy among the completed runs on every backbone. In contrast, the two baseline TTL methods vary substantially across models: both improve Qwen3.6-35B and GPT-OSS-120B but reduce accuracy on Qwen3.5-9B, and neither adds more than $2.5$ points on DeepSeek-V4-Flash. Thus, the same experience-learning approach can improve one backbone yet degrade another, consistent with the model-dependent failure profiles observed in \S\ref{sec:motivation}. The largest gain occurs on Qwen3.6-35B, the backbone examined in detail in our process-level study, where \method{} increases accuracy from 26.4\% to 45.6\%.

\subsection{Playbook Stage Ablations}
\label{sec:experiments:ablation}

We examine how the three playbook stages contribute on GPT-OSS-120B and Qwen3.6-35B by removing the learned entries from one or two sections and re-evaluating the frozen agent on the held-out split. In Table~\ref{tab:stage-ablation}, removing any single stage reduces accuracy on both backbones, with the largest single-stage decrease occurring when \textsc{Find} is removed from the Qwen3.6-35B playbook ($-3.7$ points). Removing two stages produces non-additive effects across combinations, with the largest decrease reaching $-5.4$ points on Qwen3.6-35B when both \textsc{Verify} and \textsc{Decide} are removed. Still, every ablated variant remains above \vanilla{}, indicating that the full improvement is not attributable to any single learned stage.

\section{Analysis and Discussion}
\label{sec:analysis}

Table~\ref{tab:main-results} establishes that \method improves accuracy. We now ask what those gains consist of. We apply the same six-family taxonomy used in \S\ref{sec:motivation} to all incorrect trajectories for \deepseek, \gptoss, and \qwenm.
We also pair each adapted system with its \vanilla{} agent by question identifier, separating gains into repaired questions and regressions.

\subsection{FINDER Repairs Bottlenecks with Less Outcome Churn}
\label{sec:analysis:repair}

\begin{figure}[t]
  \centering
  \begin{subfigure}{\columnwidth}
    \centering
    \includegraphics[width=0.7\columnwidth]{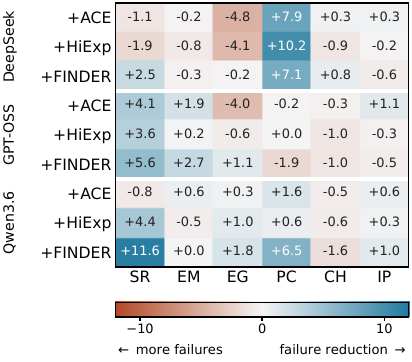}
    \vspace{-5pt}
    \caption{Change in failure-family incidence.\label{fig:process-changes:failure}}
  \end{subfigure}
  \begin{subfigure}{\columnwidth}
    \centering
    \includegraphics[width=0.7\columnwidth]{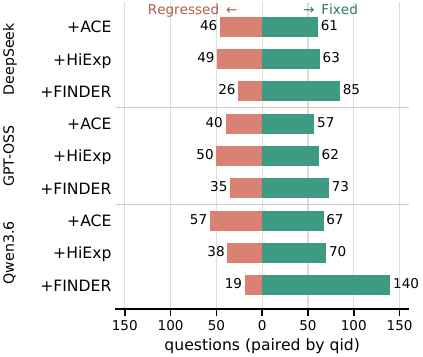}
    \vspace{-5pt}
    \caption{Paired question outcomes.\label{fig:process-changes:change}}
  \end{subfigure}
  \vspace{-20pt}
  \caption{Process changes of each adaptation method relative to the
  backbone's \vanilla{} agent. In (a), each cell is the change in one
  family's failure incidence in percentage points of the 630 held-out
  questions; positive cells denote failure reduction. The panel
  covers trajectories with taxonomy labels. In (b), questions are paired
  by task ID and missing evaluations count as incorrect: \emph{fixed}
  questions change from wrong to right, and \emph{regressed} questions
  change from right to wrong. \method{} reduces the dominant families
  without the error transfer and outcome churn seen in several baseline
  runs.}
  \Description{Top: a heatmap with rows for ACE, HiExp, and FINDER on
  DeepSeek, GPT-OSS, and Qwen3.6, and columns for six failure families,
  colored blue for failure reduction and red for added failures; the
  largest reduction is +11.6 points in search and retrieval for FINDER on
  Qwen3.6. Bottom: diverging horizontal bars of regressed versus fixed
  questions for the same nine method-backbone pairs; FINDER has the fewest
  regressions on every backbone and fixes 140 Qwen3.6 questions while
  regressing 19.}
  \label{fig:process-changes}
  \vspace{-10pt}
\end{figure}

\paragraph{Gains by category: \method{} improves each backbone's own bottlenecks.}
Figure~\ref{fig:process-changes:failure} decomposes each method's change relative to \vanilla{} by failure family; a positive value is an improvement, meaning fewer failures of that family. \method{}'s largest improvements fall on the bottlenecks identified for each backbone in \S\ref{sec:motivation}. On \qwenm{}, search and retrieval(SR) improves by $11.6$ points and process control(PC) by $6.5$ points, the two largest changes in the panel. On \deepseek{}, process control improves by $7.1$ points and search and retrieval by $2.5$ points. On \gptoss{}, search and retrieval improves by $5.6$ points and entity mapping(EM) by $2.7$ points, whereas evidence grounding(EG), its secondary bottleneck, improves by only $1.1$ points. \method therefore improves the principal retrieval bottleneck on all three annotated backbones and the secondary bottleneck on two of them.

Comparing methods across backbones sharpens this picture in two ways. First, the baselines' gains are often offset by error transfer, most often into evidence grounding(EG): on \deepseek, \ace and \hiexp improve process control(PC) by even more than \method does, but at the cost of a four-point increase in EG failures, and \ace{}'s retrieval improvement on \gptoss is matched almost one-for-one by a grounding increase. Under \method{}, no family worsens by more than two points on any backbone. Second, the baselines' effect on the dominant
family is inconsistent across backbones. \ace leaves \qwenm{}'s SR failures essentially unchanged, and both baselines slightly increase them on \deepseek{}, whereas \method improves SR across every backbone, especially where the \vanilla agent's bottleneck is largest. Taken together, the category-level view shows that generic experience can move a backbone's failure profile in ways that a net score hides, whereas \method{}'s improvements are concentrated in the target families and consistent across backbones.

\paragraph{Net gains: repair rather than balanced turnover.}
Accuracy differences alone conflate wrong-to-right repairs with right-to-wrong regressions. Figure~\ref{fig:process-changes:change} separates the two. On Qwen3.6-35B, \method{} repairs 140 questions while regressing 19, compared with 67/57 for ACE and 70/38 for HiExp. The same ordering holds on DeepSeek (85/26 versus 61/46 and 63/49) and GPT-OSS (73/35 versus 57/40 and 62/50). Equivalently, the regression-to-repair ratio is $0.14$--$0.48$ for \method{} and $0.54$--$0.85$ for the two baselines. The repaired questions come from the intended bottlenecks: search and retrieval plus process control account for 119 of Qwen's 140 repairs and 70 of DeepSeek's 85, while search and retrieval plus evidence grounding account for 58 of GPT-OSS's 73. These paired transitions support a stronger conclusion than the marginal failure counts alone: \method{}'s gains arise from repairing a concentrated set of \vanilla{} failures while preserving more of \vanilla{}'s successes.

\subsection{What the Playbook Learns}
\label{sec:analysis:experience}

\begin{table}[t]
  \centering
  \small
  \caption{Composition of the experience artifacts. Words
  are whitespace-delimited counts over the stored textual fields.
  }
  \resizebox{\columnwidth}{!}{%
  \begin{tabular}{lrrrcrc}
    \toprule
    & \multicolumn{2}{c}{\ace{}} & \multicolumn{2}{c}{\hiexp{}}
    & \multicolumn{2}{c}{\method{}} \\
    \cmidrule(lr){2-3}\cmidrule(lr){4-5}\cmidrule(lr){6-7}
    Backbone & Entries & Words & Entries & Words
    & Entries (F/V/D) & Words \\
    \midrule
    \deepseek{} & 152 & 8,647 & 31 & 4,879 & 16 (10/1/5) & 5,599 \\
    \gptoss{}     & 319 & 16,372 & 32 & 4,100 & 19 (8/3/8) & 2,703 \\
    \qwenm{}  & 124 & 13,520 & 42 & 6,146 & 12 (8/3/1) & 3,981 \\
    \qwens{}       & 116 & 12,998 & 53 & 9,420 & 11 (6/1/4) & 2,188 \\
    \bottomrule
  \end{tabular}}
  \label{tab:experience-scale}
  \vspace{-25pt}
\end{table}

Having established the outcome changes in \S\ref{sec:analysis:repair}, we next examine the learned artifacts. Table~\ref{tab:experience-scale} shows that \method retains 11--19 learned entries, compared with 116--319 for \ace and 31--53 for \hiexp{}. Its section composition also varies by backbone: \qwenm{} is \textsc{Find}-heavy (8/3/1), whereas \gptoss{} retains as many \textsc{Decide} entries as \textsc{Find} entries (8/3/8). The word counts prevent an overly simple compactness claim. \method stores less text than \ace on every backbone, but its \deepseek{} artifact is slightly larger than \hiexp's and some individual \method entries are long. Artifact size and stage organization are therefore observed properties, and the removal experiments in Table~\ref{tab:stage-ablation} provide the more direct evidence that content from all three sections contributes.

\begin{figure}[t]
  \centering
  \footnotesize
  \fbox{\begin{minipage}{0.96\columnwidth}
  \textbf{Adaptation question (qid 1223; gold answer: 30/07/1980)}%
  \par\smallskip
  A former head of state of a country that has a creole as one of its
  official languages has the same first name as a popular Disney character.
  More than half of that country's population was under 25, and the
  population growth rate was in the range of 2.2\% to 2.5\% in 2023. Can
  you give me the date of independence of this country as of 2023 in
  DD/MM/YYYY format?
  \par\medskip\hrule\medskip
  \textbf{\ace{} bullet --- \gptoss{}, [ss-00156] (verbatim)}\par\smallskip
  When a clue references a former head of state
  whose name matches a pop-culture character (e.g., ``Moana''), treat this
  as a high-impact relational filter: first compile a list of all past
  leaders of candidate entities, then intersect with the name-match
  condition before applying broader demographic or linguistic filters.
  \par\medskip\hrule\medskip
  \textbf{\method{} entry --- \gptoss{}, [F-012] (verbatim)}\par\smallskip
  \textbf{Task}: When locating an entity described by many clues, first
  isolate the most unique, low-entropy attribute (e.g., official language,
  capital, unique title) and query that alone before adding further
  filters.\par
  \textbf{Pre-search state}: No candidate identified; need an initial
  anchor that uniquely points to the target.\par
  \textbf{Good query}: \texttt{Vanuatu official language Bislama} --- A
  single highly specific clue creates a focused dense embedding that
  directly retrieves the correct entity page, providing a solid foundation
  for subsequent verification.\par
  \textbf{Bad query}: \texttt{official language Bislama country population
  growth 2.3\% 2023} --- Bundling many constraints dilutes the signal and
  can miss the target document.\par
  \textbf{Caveat}: Keep the query a single natural-language sentence
  without quotes or Boolean operators; add extra constraints only after
  the initial retrieval.
  \end{minipage}}
  \vspace{-5pt}
  \caption{Same backbone (\gptoss{}), same adaptation question, two
  methods (\ace and \method{}). Both artifacts identify the Disney-name clue, so topical
  coverage is equal; they differ in action form.}
  \Description{A boxed text excerpt showing one adaptation question about
  a country with a creole official language, followed by the verbatim ACE
  bullet and FINDER entry that GPT-OSS-120B learned from it. The ACE
  bullet is a single block of advice; the FINDER entry is split into a
  task, a pre-search state, a good query, a bad query with the reason it
  fails, and a caveat on query form.}
  \label{fig:playbook-contrast}
  \vspace{-10pt}
\end{figure}

\paragraph{A case study of the learned playbooks.}
We present a case study that reads the three audited \method playbooks directly and contrasts the entries for one \gptoss{} adaptation item with the \ace{} bullet learned from the same question. The case study is qualitative: it reports what the stored entries say and where they come from, and it does not measure which entry was invoked on which held-out question. As hypothesized in \S\ref{sec:motivation}, each entry binds a lesson to a decision state rather than naming a failure topic, and its \textsc{Find} and \textsc{Verify} entries record query contrasts tested against the deployed search engine (Figure~\ref{fig:playbook-contrast}). Although learned separately, the three playbooks converge on two themes. The first is search-engine-aware anchoring: identify the rarest discriminator in a clue bundle, such as a proper noun, a datable event, a verbatim quote, or a scoreline, and use it to anchor the query. As one \deepseek entry puts it, the fix is ``to change \emph{which} attributes are in the query \ldots, not to rephrase the same ones.'' The second is commit discipline: once a candidate matches the rarest constraints, run at most one confirming query and then answer. In the words of another entry, ``recurring identical docids across queries is EVIDENCE the candidate is right, not a prompt to search more.'' One \qwenm{} entry consolidates four adaptation episodes in which the agent ended without an answer after 20--30 verification searches. These two themes address search and retrieval (SR) and process control (PC), the two dominant \vanilla{} failure families in \S\ref{sec:motivation}.

Beyond these shared themes, each playbook also retains emphases specific to its backbone's failure profile. Four of \deepseek{}'s five \textsc{Decide} entries standardize the exact answer form: strip a corporate suffix, prefer a formal name to a nickname, use the canonical transliteration, or return the bare proper noun. Each was distilled from an episode in which retrieval succeeded but the final answer failed the judge. By contrast, three of \gptoss{}'s eight \textsc{Decide} entries require grounding the answer in specific evidence, such as the exact values in a league table, an event date rather than its announcement date, or a value reported in the cited paper. This emphasis reflects the model's evidence-grounding bottleneck. \qwenm instead places eight of its twelve entries in \textsc{Find}. A single adaptation question, which asks for the royal family into which a described person's spouse was born, makes this specificity concrete. \gptoss learned to combine the spouse's rare attributes in one query, whereas \qwenm learned to pivot to the associate with the more distinctive attributes. \deepseek found the spouse but kept searching the original clues and then over-copied the answer, so its playbook records both a \textsc{Verify} pivot and a \textsc{Decide} rule that commits ``Nwoko'' rather than ``Nwoko Royal Family.'' The same question thus yields guidance for a different failure point on each backbone.

The case study also reveals a cross-method difference in actionability. Figure~\ref{fig:playbook-contrast} compares what \ace and \method store from the same \gptoss adaptation item. The \ace bullet turns the clue about a former head of state sharing a Disney character's first name into a database-style plan: compile all past leaders and then intersect that list with the name constraint. This plan identifies a useful clue but does not specify how to realize either step with the agent's single free-text search action. \method instead stores query strategies executed against the deployed search engine. The entry shown in the figure first isolates a low-entropy attribute with the query \texttt{Vanuatu official language Bislama} and adds other constraints only after retrieving a candidate, and a companion entry for the same item covers the complementary case of combining jointly specific constraints in one natural-language query. This example instantiates the coverage--outcome mismatch from \S\ref{sec:motivation} at the level of a single lesson: recognizing a relevant clue does not by itself yield guidance that the deployed agent can execute directly.

\subsection{Search Productivity, Not Only Search Volume}
\label{sec:analysis:efficiency}

\begin{figure}[t]
  \centering
  \includegraphics[width=0.65\columnwidth]{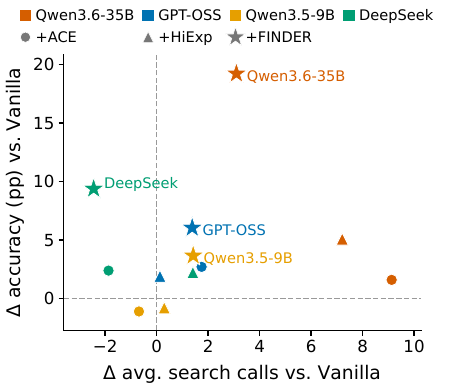}
\vspace{-10pt}
  \caption{Accuracy gain versus the change in average search calls, relative
  to the same backbone's \vanilla{} agent.}
  \Description{A scatter plot of accuracy change in percentage points
  against change in average search calls, both relative to Vanilla, with
  colors for four backbones and marker shapes for ACE, HiExp, and FINDER.
  FINDER stars lie highest for every backbone: Qwen3.6-35B near 19 points
  with about 3 extra searches, and DeepSeek near 9 points with about 2.4
  fewer searches. ACE and HiExp points mostly lie below 5 points, and two
  Qwen3.6-35B baselines use 7 to 9 extra searches.}
  \label{fig:efficiency}
  \vspace{-15pt}
\end{figure}

Figure~\ref{fig:efficiency} shows that additional search volume does not explain the main result. \method{} adds 3.10 average searches on Qwen3.6-35B for a 19.21 accuracy gain and 1.38 searches on GPT-OSS for a 6.04 accuracy gain. On DeepSeek, it gains 9.37 points while reducing search by 2.44 calls. Qwen3.5-9B is the least favorable trade-off, adding 1.42 calls for a 3.65 accuracy gain. By comparison, \ace and \hiexp add 9.13 and 7.21 searches on Qwen3.6-35B for substantially smaller improvements.

\begin{table}[t]
  \centering
  \small
  \caption{Paired search behavior on questions that \method{} newly solves.
  Deltas are \method{} minus \vanilla{}; Novel is the novel-document ratio, and
  Hit limit is the percentage of these questions whose trajectory ends by
  exhausting the search or turn budget under each system.}
  \vspace{-5pt}
  \resizebox{\columnwidth}{!}{%
  \begin{tabular}{lrrrrrr}
    \toprule
    & & & & & \multicolumn{2}{c}{Hit limit (\%)} \\
    \cmidrule(lr){6-7}
    Backbone & Fixed & $\Delta$Search & $\Delta$Repeat
    & $\Delta$Novel & Vanilla & \method{} \\
    \midrule
    DeepSeek-V4-Flash & 85  & $-5.75$ & $+0.00$ & $+0.056$ & 9.4 & 2.4 \\
    GPT-OSS     & 73  & $+0.38$ & $+0.08$ & $-0.019$ & 0.0 & 0.0 \\
    Qwen3.6-35B-A3B & 140 & $+0.01$ & $-7.46$ & $+0.019$ & 17.1 & 0.0 \\
    Qwen3.5-9B  & 57  & $-6.28$ & $-0.04$ & $+0.025$ & 0.0 & 0.0 \\
    \bottomrule
  \end{tabular}}
  \label{tab:fixed-efficiency}
  \vspace{-10pt}
\end{table}

The paired audit in Table~\ref{tab:fixed-efficiency} clarifies how a system can improve search without uniformly reducing the number of calls. On the 140 Qwen3.6-35B questions newly solved by \method{}, the mean call count is virtually unchanged, but redundant queries fall by 7.46 per question, the novel-document ratio rises, and limit hits fall from 17.1\% to zero. Newly solved DeepSeek and Qwen3.5-9B questions use 5.75 and 6.28 fewer searches on average. GPT-OSS is the exception: its repaired set uses 0.38 more searches and has slightly worse redundancy and novelty. The consistent result is thus not that \method{} always searches less, but that its largest repair sets are not obtained by indiscriminately increasing search volume.

\section{Conclusion}
\label{sec:conclusion}

This paper studied how context-level test-time learning changes in a deep-research agent. Applying one failure taxonomy to both the agent's trajectories and the experience it learns, we found that failures concentrate in a few model-specific bottlenecks, that generic experience raises accuracy without reliably repairing them, and that the learned artifacts already cover the relevant families. This coverage--outcome mismatch motivated \method{}, which binds each lesson to the stage at which the agent breaks down, grounds query lessons in the deployed retriever, and curates them under a bounded budget. Across four frozen backbones, \method{} improves accuracy by 3.7 to 19.2 points, and the same diagnostic shows that its gains repair each backbone's own bottlenecks with few regressions and are not explained by additional search. The playbooks suggest a broader lesson: much of what a search agent must learn at test time is how its own search tool behaves.

\section*{Limitations}

Our process-level evidence rests on a single LLM annotator, so the family shares are measurements under one protocol rather than ground truth.
Adaptation requires gold answers on the adaptation split to judge rollouts, so the setting is supervised rather than fully online, and grounding is tool-assisted rather than certified, since a lesson is not rejected when it omits the query it was asked to run. Results are from one run per configuration; analysis covers three of four backbones. The fixed playbook skeleton is retained in the stage ablations, so they bound each learned section's contribution without separating it from the skeleton's static guidance.

\appendix

\section{Experiment Settings}
\label{sec:appendix:settings}

All methods share one evaluation stack. Search is dense retrieval with Qwen3-Embedding-8B over the official BrowseComp-Plus corpus of roughly 100K documents; each search call returns the top five documents, truncated to 512 tokens. Agents decode with temperature 1.0, top-$p$ 0.95, up to 8{,}192 response tokens, and thinking enabled. \method{}'s \reflector{} and \optimizer{} are instantiated by the same frozen backbone as the agent. Each adaptation step processes $B{=}4$ questions with $G{=}4$ rollouts per question, so the 100-question stream yields 25 steps and 400 rollouts, matching the budget used to construct \hiexp{}'s experience.
\method{}'s reported results use the playbook at the end of adaptation.

\section{Baseline Reproduction}
\label{sec:appendix:baselines}

\paragraph{Our instantiation of \ace{}}
We use the official \ace{} reflector and curator with unmodified prompts and replace its single-call generator with the shared ReAct search agent, so all methods share the same loop, tools, and prompts. The generator, reflector, and curator use the backbone under adaptation. Adaptation runs offline for one epoch over the 100 adaptation questions, with up to three reflect-and-regenerate rounds on incorrect answers and curation after every question. The reflector receives the question, the gold answer, and the agent's final response, reflecting on the answer rather than the search trajectory. At evaluation, the complete frozen playbook is appended to every rollout's system prompt.

\paragraph{Our prompt-based implementation of \hiexp{}.}
\hiexp{} has no public code release, so we implement its training-free variant from the paper's equations and prompts, used verbatim. Experience is built from four judged rollouts for each of the 100 adaptation questions. Each question with both a successful and a failed rollout contributes one contrastive pair, from which the backbone distills a case-level E1 entry. E1 titles are embedded with multilingual-e5-base and clustered with Ward linkage (distance threshold 0.6, singletons dropped), and each cluster is summarized into a strategy-level E2 entry. At inference, the five E2 entries most similar to the question enter the system prompt, and E1 entries whose titles match a new search query above cosine similarity 0.8 (top 3) are injected into the tool response. The store is frozen during evaluation.

\section{Failure Taxonomy and Annotation Protocol}
\label{sec:appendix:taxonomy}

Trajectories are annotated by DeepSeek-V4-Pro with reasoning enabled and default API sampling. For each trajectory judged incorrect, the annotator receives the question, the gold answer, and the linearized trajectory and is instructed to select the single family most directly responsible for the wrong answer. Experience entries are annotated with the same annotator and codes, one entry at a time without playbook context (an \ace{} bullet, a \method{} entry, or the fields of a \hiexp{} E1/E2 record), with the instruction to select the family the entry is most directly intended to prevent or remedy.

\begin{acks}
We thank the anonymous reviewers for their feedback.
This work was supported in part by the Natural Sciences and Engineering Research Council of Canada (NSERC) under funding reference number RGPIN-2024-04909.
This work is enabled in part by support provided by Compute Ontario (computeontario.ca) and the Digital Research Alliance of Canada (alliancecan.ca).
\end{acks}

\section*{Ethical Considerations}

This work uses only the public BrowseComp-Plus benchmark, whose questions and fixed, human-verified corpus concern publicly documented entities, and all search runs against that corpus rather than the live web. No human subjects or crowd workers were involved, and the failure annotations come from an automated annotator whose limitations we report. The learned playbooks are plain text that can be inspected and edited, which supports oversight, but because lessons are distilled from benchmark questions, some entries retain the names of public figures that appear in the corpus. We therefore recommend that deployments of context-level test-time learning abstract such identifiers into entity types, as our curation prompts already encourage, and review accumulated experience before reuse. More broadly, improving the ability of agents to locate hard-to-find information about individuals carries a dual-use risk for profiling or surveillance. Our study is confined to a closed corpus and does not release new retrieval capabilities, but the same process-level diagnostics could help practitioners audit what a deployed agent has learned to search for. Finally, the weight-frozen setting avoids model training and its associated compute cost.

\bibliographystyle{ACM-Reference-Format}
\bibliography{sample-base}

\end{document}